\documentclass{article}
\usepackage{arxiv}

\usepackage{booktabs} 
\usepackage{CJKutf8}
\usepackage{algpseudocode}
\usepackage {setspace}
\usepackage{mathtools}
\usepackage{comment}
\usepackage{booktabs} 
\usepackage{url}
\usepackage{helvet}  
\usepackage{courier}  
\usepackage{graphicx}  
\usepackage{subfigure}
\usepackage{courier}
\usepackage{times}
\usepackage{color, colortbl}
\usepackage{amssymb,amsmath}
\usepackage{booktabs}
\usepackage{arydshln}
\usepackage{enumitem,kantlipsum}
\usepackage[utf8]{inputenc} 
\usepackage[T1]{fontenc}    
\usepackage{hyperref}       
\usepackage{amsfonts}       
\usepackage{nicefrac}       
\usepackage{microtype}      
\usepackage{lipsum}

\usepackage{tikz}
\usetikzlibrary{bayesnet}
\allowdisplaybreaks
\usepackage{times}
\usepackage{helvet}
\usepackage{courier}
\usepackage{url}
\usepackage{graphicx}
\title{Empirical Evaluation of Multi-task Learning in Deep Neural Networks for Natural Language Processing}

\renewcommand\[{\begin{equation}}
\renewcommand\]{\end{equation}}

\author{\rm{Jianquan Li}\textsuperscript{1}
\and Xiaokang Liu\textsuperscript{1}
\and Wenpeng Yin\textsuperscript{2}
\and Min Yang\textsuperscript{3}
\and Liqun Ma\textsuperscript{1}
\and Yaohong Jin \textsuperscript{1}
}

\date{\small{\textsuperscript{1}Beijing Ultrapower Software Co.,Ltd. \\ 
\textsuperscript{2}Department of Computer and Information Science, University of Pennsylvania \\
\textsuperscript{3}Shenzhen Institutes of Advanced Technology, Chinese Academy of Sciences \\ 
}
\small{
\textsuperscript{1}\{lijianquan2,liuxiaokang1,maliqun, jinyaohong\}@ultrapower.com.cn, 
\textsuperscript{2} min.yang@siat.ac.cn, 
\textsuperscript{3}wenpeng@cis.lmu.de}
}
\makeatother
\begin{document}
\maketitle
\begin{abstract}
Multi-Task Learning (MTL) aims at boosting the overall performance of each individual task by leveraging useful information contained in multiple related tasks.
It has shown great success in natural language processing (NLP).
Currently, a number of MTL architectures and learning mechanisms have been proposed for various NLP tasks, including exploring linguistic hierarchies, orthogonality constraints, adversarial learning, gate mechanism, and label embedding.  
However, there is no systematic exploration and comparison of different MTL architectures and learning mechanisms for their strong performance in-depth.
In this paper, we conduct a thorough examination of five typical MTL methods with deep learning architectures for a broad range of representative NLP tasks. 
Our primary goal is to understand the merits and demerits of existing MTL methods in NLP tasks, thus devising new hybrid architectures intended to combine their strengths. Following the empirical evaluation, we offer our insights and conclusions regarding the MTL methods we have considered. 
\end{abstract}

\section{Introduction}
Multi-Task Learning (MTL) with deep neural networks have recently shown promising results in many NLP tasks, especially when there is no sufficient training data.  The main idea of MTL is to leverage useful information contained in multiple related tasks to improve the generalization performance of all the tasks \cite{zhang2017survey}.  Generally,  existing MTL methods with deep neural networks can be divided into two primary categories: \textit{hard parameter sharing} and \textit{soft parameter sharing}. 
Hard parameters sharing is the most common used MTL framework, which shares the hidden layers among all tasks but each task has its specific output layer \cite{liu2019multi,liu2016recurrent,yang2016multi}). 
While for soft parameters sharing, each task has its own neural network architecture with specific parameters.  Information is passed by skip connections between the architecture for each task \cite{xiao2018gated,misra2016cross}. 
In particular, we tease apart the typical hard and soft parameter sharing MTL methods in deep neural networks into five categories:  gate mechanism \cite{xiao2018gated,misra2016cross},  exploring linguistic hierarchies \cite{sogaard2016deep,hashimoto2016joint}, adversarial learning \cite{liu2017adversarial}, orthogonality constraint \cite{bousmalis2016domain,salzmann2010factorized}, and label embedding \cite{augenstein2018multi}.  We describe these five MTL methods in detail in Section 5. 

Most previous MTL methods have declared state-of-the-art results for specific tasks. 
However, they are carefully designed and evaluated on selected (often one or two) datasets that can demonstrate the superiority of the model.  This raises the scientific questions of whether the successes of these MTL approaches are confined to specific tasks and how much performance gain is due to certain architecture designs rather than hyper-parameter optimizations.

To address these questions and understand different MTL architectures in-depth, we conduct systematically empirical evaluation of the widely used MTL architectures with deep neural networks on a broad range of NLP tasks, including text classification, semantic textual similarity, natural language inference,  question answering,  part-of-speech tagging, and chunking. 
By identifying the essential MTL architecture designs that contribute significantly to the success of MTL, we propose a hybrid architecture built with different components borrowed from previous MTL methods. For fair comparison, we merely retain the base structure of each MTL architecture and remove the training tricks in the original works. 

To the best of our knowledge, we are the first to conduct a thorough examination of existing MTL methods on various NLP tasks and present a series of findings with quantitative measurements and in-depth analysis:
\begin{itemize}
\item Multi-task learning methods have significant improvements over the single models on all the datasets. In addition, we observe from the combinatorial experiments that the improvement of different multi-task learning methods cannot be superimposed.  
\item Using linguistic hierarchical information performs better than other individual MTL methods. Surprisingly, combining all the auxiliary basic tasks cannot achieve the best results. 
\item Combining linguistic hierarchies, gate mechanism, and label embedding methods  can achieve best results on most of the datasets. 
\end{itemize}

The rest of this manuscript is structured as follows. Section \ref{sec:related-work} reviews and discusses the related references. Section \ref{sec:problem-definition} provides the problem definition of multi-task learning in typical natural language processing tasks.  The basic MTL architecture is described in Section \ref{sec:basic}. 
Section \ref{sec:different-MTL} introduces five widely used MTL methods, including exploring linguistic hierarchies, orthogonality constraints, adversarial learning, gate mechanism, and label embedding.  
Section \ref{sec:setup} describes the experimental setup. Section \ref{sec:result} demonstrates and analyzes the experimental results. Section \ref{sec:conclusion} concludes this manuscript. 

\section{Related work}
\label{sec:related-work}
Multi-task learning with deep neural networks has gained increasing attention within NLP community over the past decades. \cite{zhang2017survey} and \cite{ruder2017overview}  described most of the existing techniques for multi-task learning in deep neural networks. Generally, existing MTL methods can be categorised as \textit{soft parameter sharing} \cite{xiao2018gated,misra2016cross} and \textit{hard parameter sharing} \cite{liu2019multi,yang2016multi}. 


For soft parameter sharing, \cite{duong2015low} use the L2 distance between (selective parts of) the main and auxiliary models to regularize parameters. Analogously, \cite{yang2016trace} use the trace norm for regularization.
However, regularization can not autonomously choose which information to share. 
So, \cite{misra2016cross} and \cite{xiao2018gated} respectively proposed \textit{cross-stitch unit} and \textit{gate sharing unite} to learn shared representations in neural network. 

Hard parameters sharing without extra learning technology of MTL is widely used.
\cite{liu2016recurrent} and \cite{yang2016multi} utilize simple hard parameters sharing structure to jointly learning multi-domain text classification tasks and different sequence-tagging tasks across languages respectively.
And \cite{liu2019multi}  use the pre-trained BERT as shared layers and obtain new state-of-the-art results on ten NLP tasks. 
Visibly, the feature learned by hard parameter sharing divide into private and share feature space.
\cite{salzmann2010factorized}  and \cite{bousmalis2016domain}  argued that private feature space and share feature space may be mixed with each other and results in the reduction of model's performance. 
So, they proposed orthogonality constraints to alleviate this issue.
\cite{liu2017adversarial} proposed adversarial training framework for the same sake. 
In addition, \cite{sogaard2016deep} and \cite{hashimoto2016joint} introduced the information of linguistic hierarchies into share feature space by predicting increasingly complex NLP tasks at successively deeper share layers.

The methods mentioned above do not consider the information of tasks' label, so \cite{augenstein2018multi} proposes to induce a joint label embedding space using a label embedding layer that can model relationships between different NLP task and discovered it is effective for MTL.

\section{Problem Definition}
\label{sec:problem-definition}
We assume that there are $T$ tasks for multi-task learning, denoted as $\mathcal T_1, \mathcal T_2, \dots,\mathcal T_T$. The training data of each task is represented as $\mathcal D^{\mathcal T_k}$, where $k\in{\{1,2,\dots,T\}}$. Let $\mathcal{L}_{\mathcal T_k}(\mathcal D^{\mathcal T_k}, \Omega)$ stands for the loss function of task $\mathcal T_k$, where $\Omega$ donates the total parameters in MTL model. The aim of MTL is finding $\Omega^{*}$ which accords with following equation:
\begin{equation}
  \Omega^{*}=\mathop{\arg\min}_{\Omega}\sum_{k=1}^{T}\lambda^{\mathcal T_k}\mathcal{L}_{\mathcal T_k}(\mathcal D^{\mathcal T_k}, \Omega)
  \label{eq_1}
\end{equation}
where $\lambda^{\mathcal T_k}$ is the weight for task $\mathcal T_k$.
  
The instance of train data in $\mathcal D^{\mathcal T_k}$ is noted as $(\mathbf{x}^{\mathcal T_k}, \mathbf{y}^{\mathcal T_k})$, where  $\mathbf{x}^{\mathcal T_k} = (\mathbf{x}^{\mathcal T_k}_1, \dots , \mathbf{x}^{\mathcal T_k}_{l_{k}})$, $\mathbf{x}^{\mathcal T_k}_i \in \mathbb{R}^e$ is a $e$-dimensional word embedding of $i$-th word in the $\mathcal T_k$-th task's sentence and  $\mathbf{y}^{\mathcal T_k}=\{\mathbf{y}^{\mathcal T_k}_1, \mathbf{y}^{\mathcal T_k}_2,\dots,\mathbf{y}^{\mathcal T_k}_{N^{\mathcal T_k}}\}$ is the corresponding ground-truth label , $N^{\mathcal T_k}$ is the number of  class categories for task $\mathcal T_k$, $l_{k}$ is the length of the sentence.

\section{Basic MTL Architecture}
\label{sec:basic}
\begin{figure}[htbp] 
    \centering 
    \includegraphics[width = 0.35\columnwidth]{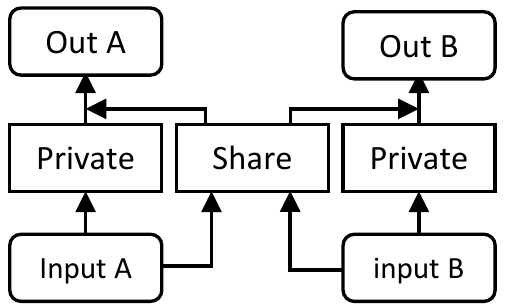} 
    \caption{The overview of the basic MTL framework. The shared extractor is composed of several BiLSTM layers that are shared by all tasks. The private extractor aims at extracting the task-specific features, and we use a BiLSTM layer to learn the sentence representations. } 
    \label{fig_base}
  \end{figure}

In order to experiment easily with different MTL methods, we define an MTL framework which contains several combinable and nestable blocks, as illustrated in Figure \ref{fig_base}.
 Next, we will elaborate on this basic MTL framework in detail. 
 
\subsection{Shared Part}
  The whole MTL framework can divide into shared part and private part roughly. 
  Shared part is composed of M-layer BiLSTMs\cite{hochreiter1997long} that are shared by all tasks.  
  Formally, the hidden state of each BiLSTM layer at step $i$ is calculated by:
  \begin{equation}
    \mathbf{s}_{i}^{\mathcal T_k, n}={\rm BiLSTM}(\mathbf{\hat{x}}^{\mathcal T_k}_i, \mathbf{s}_{i-1}^{\mathcal T_k, n})
    \label{eq_2}
  \end{equation}
  where 
  $\mathbf{s}_{i}^{\mathcal T_k, n}\in \mathbb{R}^{2l_h}$ is $i$-th output vector in $n$-th layer of task $\mathcal T_k$'s input sentence, $n\in{\{1,2,\dots,M\}}$. And $\mathbf{\hat{x}}^{\mathcal T_k}_i=[\mathbf{x}^{\mathcal T_k}_i;\mathbf{s}_{i}^{\mathcal T_k, 1};\mathbf{s}_{i}^{\mathcal T_k, 2};\dots;\mathbf{s}_{i}^{\mathcal T_k, n-1}]$. 
  In our experiments, $M=3$.
  
\subsection{Private Part}
Each task has its own task-specific structures, each of which is composed of private feature extractor, pooling layer, and output layer.

\paragraph{Private Feature Extractor} Private Feature Extractor aims at extracting the task-specific features of each task's data.
Here, it is a 1-layer BiLSTM for single sentence modeling. The hidden state of each BiLSTM layer at step $i$ is calculated by:
  \begin{equation}
    \mathbf{p}_{i}^{\mathcal T_k}={\rm BiLSTM}(\mathbf{x}^{\mathcal T_k}_i, \mathbf{p}_{i-1}^{\mathcal T_k})
    \label{eq_3}
  \end{equation}
  where $\mathbf{p}_{i}^{\mathcal T_k}\in \mathbb{R}^{2l_h}$ is the $i$-th private hidden state in the given sentence for task $\mathcal T_k$.
 And it is a 2-layer BiLSTMs with soft alignment attention between BiLSTMs are adopted for sentence pair modeling.

 \paragraph{Task Feature Extractor}
We employ another BiLSTM to encode the feature from both private part and share part if exist. We define the input as $\widetilde{\mathbf{p}}_{i}^{\mathcal T_k}$ and the output as $\hat{\mathbf{p}}_{i}^{\mathcal T_k}$, and the content of $\widetilde{\mathbf{p}}_{i}^{\mathcal T_k}$ is depended on the method we choose. When we adopt shared part, $\widetilde{\mathbf{p}}_{i}^{\mathcal T_k}= [\mathbf{p}^{\mathcal T_k}_i;\mathbf{s}^{\mathcal T_k, 1}_i;\mathbf{s}^{\mathcal T_k, 2}_i;\mathbf{s}^{\mathcal T_k, 3}_i]$. When we remove shared part, $\widetilde{\mathbf{p}}_{i}^{\mathcal T_k}= \mathbf{p}^{\mathcal T_k}_i$. The hidden state of each BiLSTM layer at step $i$ is calculated by:
  \begin{equation}
    \hat{\mathbf{p}_{i}}^{\mathcal T_k}={\rm BiLSTM}(\widetilde{\mathbf{p}}^{\mathcal T_k}_i, \hat{\mathbf{p}}_{i-1}^{\mathcal T_k})
    \label{eq_3}
  \end{equation}


\paragraph{Pooling} We  convert the learned feature vectors $\hat{\mathbf{p}}_{i}^{\mathcal T_k}$ to a fixed-length vector using pooling operation.
Here, we adopt max and mean pooling operations \cite{chen2016enhanced}:
\begin{gather}
    \mathbf{v}_{mean}^{\mathcal T_k}=\frac{1}{l_{sen}}\sum_{i=1}^{l_{sen}}\hat{\mathbf{p}}_{i}^{\mathcal T_k}\\
    \mathbf{v}_{max}^{\mathcal T_k}=\max_{i=1}^{l_{sen}}\hat{\mathbf{p}}_{i}^{\mathcal T_k} \label{eq_4}\\
  \mathbf{v}^{\mathcal T_k}=[\mathbf{v}_{mean}^{\mathcal T_k};\mathbf{v}_{max}^{\mathcal T_k}]
  \label{eq_5}
\end{gather}
  where $\mathbf{v}^{\mathcal T_k}\in \mathbb{R}^{4l_h}$ is the final sentence embedding for task $\mathcal T_k$-th.

\paragraph{Output Layer} Output layer is a fully-connected neural network with softmax, which outputs the probability distribution of corresponding labels:
\begin{equation}
    \mathbf{\hat{y}^{\mathcal T_k}} = softmax(\mathbf{W}_k\mathbf{v}^{\mathcal T_k}+b_k)
    \label{eq_6}
\end{equation}
where $\mathbf{W}_k$ is a learnable parameter for $\mathcal T_k$, $\mathbf{\hat{y}^{\mathcal T_k}}\in \mathbb{R}^{N^{\mathcal T_k}}$.
Finally, we can define the loss $\mathcal{L}_{task}$ as:
  \begin{equation}
    \mathcal{L}_{task}=\sum_{t=1}^{T}\lambda^{\mathcal T_k}\mathcal{L}_{\mathcal T_k}
    \label{eq_7}
  \end{equation}
where $\mathcal{L}_{\mathcal T_k}$ is the loss function of task ${\mathcal T_k}$ and $\lambda^{\mathcal T_k}$ is the weight for task $\mathcal T_k$.

\section{Different MTL Mechanisms}
\label{sec:different-MTL}

In this section, we describe the five MTL methods in detail.

\subsection{Exploring Linguistic Hierarchies (ELH).}
\begin{figure}[htbp] 
    \centering 
    \includegraphics[width = 0.45\columnwidth]{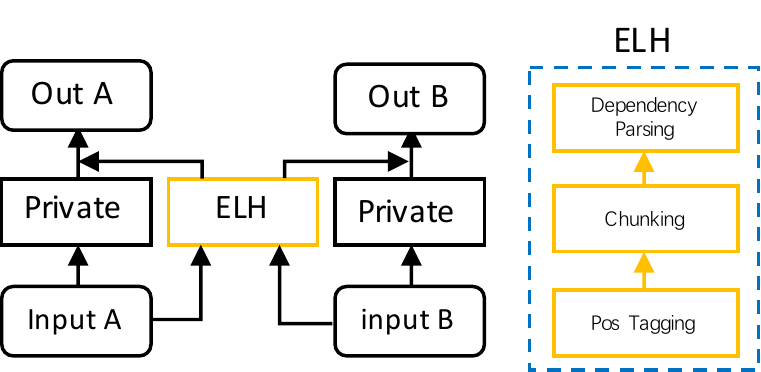} 
    \caption{Integrating the ELH method into the basic MLT framework.} 
    \label{fig_ELH}
  \end{figure}

\cite{sogaard2016deep}  and \cite{hashimoto2016joint}
 introduced the joint many-task model, which considers the linguistic hierarchies by predicting increasingly complex NLP tasks at successively deeper layers. In this study, we assume that the ELH method uses three BiLSTM layers to handle three different fundamental tasks in the order of POS tagging, chunking, dependency parsing by considering linguistic hierarchies, rather than handling different tasks in the same abstractive layer. Figure \ref{fig_ELH} illustrates the overview of integrating the ELH method into the basic MLT framework. 
Note that this method should be used together with simple hard parameters sharing method. 
In the experiments,  we define the final loss function as the linear combination of the loss functions of POS tagging, chunking,  dependency parsing, and the primary task: 
\begin{equation}
    \mathcal{L}_{LH}=\mathcal{L}_{task}+\mathcal{L}_{pos}+\mathcal{L}_{chunk}+\mathcal{L}_{parse}
    \label{eq_8}
\end{equation}

\subsection{Orthogonality Constraints (OC)}
\begin{figure}[htbp] 
    \centering 
    \includegraphics[width = 0.4\columnwidth]{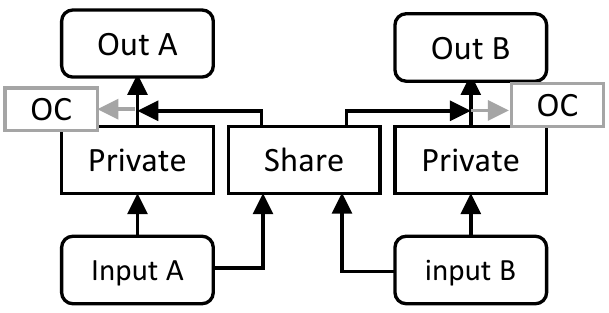} 
    \caption{Integrating the OC method into the basic MLT framework.} 
    \label{fig_OC}
  \end{figure}

\cite{bousmalis2016domain}  proposed to factorize the latent space into shared and private spaces by introducing orthogonality constraints (OC), which penalize redundant latent representations. The orthogonality constraints are encoded as minimizing the Frobenius norm of the inner product between the private and shared representations. Figure \ref{fig_OC} illustrates the overview of integrating the OC method into the basic MLT framework. 
Formally, similar to \cite{liu2017adversarial}, the orthogonality constraints are implemented by minimizing: 
\begin{equation}
    \mathcal{L}_{OC}= \lambda_{OC}\sum^T_{i=1}\sum^M_{n=1}|| (\mathbf{S}^{\mathcal T_i, n})^\top \mathbf{P}^{\mathcal T_i}||^2_F
    \label{eq_9}
\end{equation}
where, $M$ is the number of LSTM layers in shared part. $\lambda_{OC}$ is loss weight. $||\cdot||_F$ is the Frobenius norm. $\mathbf{S}^{\mathcal T_i, n}$
and $\mathbf{P}^{\mathcal T_i}$ are two metrics, whose rows are the output of shared part and private feature extractor of a input sentence.

\subsection{Adversarial Learning (AL)}
\begin{figure}[htbp] 
    \centering 
    \includegraphics[width = 0.35\columnwidth]{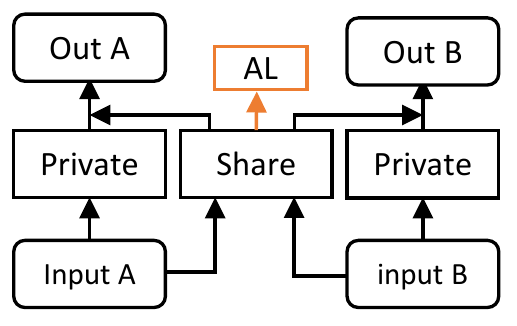} 
    \caption{Integrating the AL method into the basic MLT framework.} 
    \label{fig_AL}
  \end{figure}

  To prevent the shared and private latent feature spaces from interfering with each other, \cite{liu2017adversarial}  proposed an adversarial learning (AL) framework for multi-task learning, in which the adversarial training is used to encourage the shared features merely to contain task-invariant information. Specifically, an extra task adversarial loss $\mathcal{L}_{adv}$ is employed to encourage the model to produce shared features such that a task discriminator cannot reliably predict the task based on these features.  
  Figure \ref{fig_AL} illustrates the overview of integrating the AL method into the basic MLT framework. 
  Formally, we use a discriminator $D$ to output a probability distribution of based on the shared sentence representations, which estimates which task the encoded sentence comes from:
  \begin{equation}
    D(\mathbf{s}_s^{\mathcal T_k}, \theta_D)=softmax(\mathbf{W}_D\mathbf{s}^{\mathcal T_k}+\mathbf{b}_D)
    \label{eq_10}
  \end{equation}
  where $\mathbf{W}_D$ is a learnable parameter, $\mathbf{b}_D \in \mathbb{R}^T$ is a bias term,  $\theta_D$ denotes the parameters in discriminator.

Given a sentence, the adversarial loss is proposed to force the shared BiLSTMs to generate a representation to mislead the task discriminator $D$.
Following \cite{liu2017adversarial}, the adversarial loss is extended to multi-class form, which allows the model can be trained together with multiple tasks:
 \begin{equation}
    \mathcal{L}_{adv}=\lambda_{adv}\min_{\theta_S}(\lambda \max_{\theta_D}(\sum_{k=1}^T d^{\mathcal T_k}\log[D(\mathbf{s}^{\mathcal T_k,3})]))
    \label{eq_11}
\end{equation}
  where $\lambda_{adv}$ denotes the adversarial loss weight, $d^{\mathcal T_k}$ denotes the ground-truth label indicating the type of the current task, $\theta_S$ denotes the parameters in shared part.
  
\subsection{Gate Mechanism (Gate)}
\begin{figure}[htbp] 
    \centering 
    \includegraphics[width = 0.65\columnwidth]{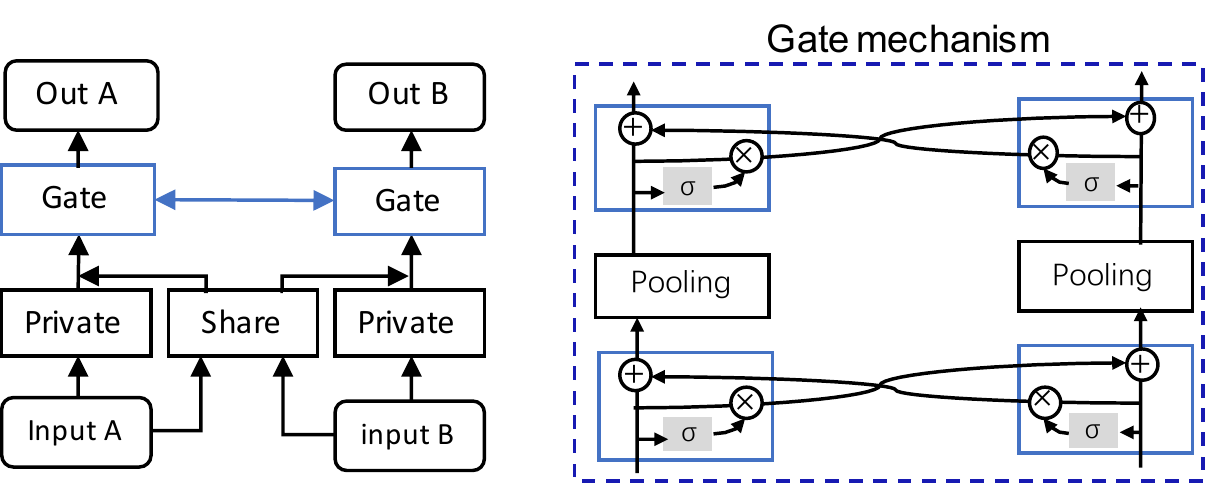} 
    \caption{Integrating the gate mechanism into the basic MLT framework.} 
    \label{fig_Gate}
  \end{figure}

  For soft parameters sharing, adding weighted futures learned by other tasks is a common way to share information among tasks \cite{xiao2018gated,misra2016cross,ruder2017learning}, in which the weights of the features are learnable parameters.  In this paper, we conclude these different weight learning mechanisms as gate mechanism. Figure \ref{fig_Gate} illustrates the overview of integrating the gate mechanism into the basic MLT framework. 

  When investigating the effectiveness of the gate mechanism, we remove the shared part from the overall MTL framework, since the gate mechanism can learn to filter the feature flows between tasks and reduce the interference. 
  Without loss of generality, we only consider two tasks: $\mathcal T_m$ and $\mathcal T_n$ in our multi-task learning framework for the sake of illustrating the gate mechanism concisely. 
  We can generalize the model to the case of multiple tasks easily.
  As shown in Figure \ref{fig_base}, each task has two gate units that learn the weights of the two kinds of features:
  \begin{gather}
    \mathbf{g}^{\mathcal T_k}_1=\sigma(\mathbf{W}_{g_1}\hat{\mathbf{p}}^{\mathcal T_k} + \mathbf{b}_{g_1}) \label{eq_12}\\
    \mathbf{g}^{\mathcal T_k}_2=\sigma(\mathbf{W}_{g_2}\mathbf{v}^{\mathcal T_k} + \mathbf{b}_{g_2})
    \label{eq_13}
  \end{gather}
where $\hat{\mathbf{p}}^{\mathcal T_m}$ and $\mathbf{v}^{\mathcal T_m}$ can be updated by:
  \begin{gather}
    \hat{\mathbf{p}}^{\mathcal T_m}=\mathbf{g}^{\mathcal T_n}_1 \odot \hat{\mathbf{p}}^{\mathcal T_n}+\hat{\mathbf{p}}^{\mathcal T_m} \label{eq_14}\\
    \mathbf{v}^{\mathcal T_m}=\mathbf{g}^{\mathcal T_n}_1 \odot \mathbf{v}^{\mathcal T_n}+\mathbf{v}^{\mathcal T_m}
    \label{eq_15}
  \end{gather}
  where $\odot$ denotes element-wise multiplication.

\subsection{Label Embedding (LE)}
\begin{figure}[htbp] 
    \centering 
    \includegraphics[width = 0.35\columnwidth]{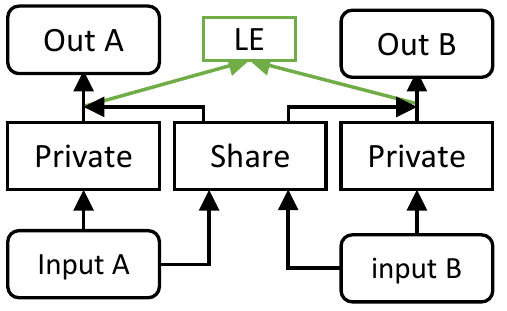} 
    \caption{Integrating the label embedding method into the basic MLT framework.} 
    \label{fig_LE}
  \end{figure}

For many NLP tasks, disparate label sets are weakly correlated, \cite{augenstein2018multi}  thus propose to induce a joint label embedding space using a label embedding layer that can model relationships between different NLP tasks. Figure \ref{fig_LE} illustrates the overview of integrating the label embedding into the basic MLT framework. 
In particular, the output of each task $\mathcal T_k$ and the loss $\mathcal{L}_{task}$ are respectively calculated by:
\begin{gather}
    \mathbf{\hat{z}}^{\mathcal T_k} = softmax(\mathbf{W}_{L}\mathbf{v}^{\mathcal T_k}+\mathbf{b}_{L}) \label{eq_16} \\
    \mathcal{L}_{task}=\sum_{k=1}^{T}\lambda^{\mathcal T_k}\sum_{i=1}^N\mathbf{z}^{\mathcal T_k}_i\log(\mathbf{\hat{z}}^{\mathcal T_k}_i)
    \label{eq_17}
\end{gather}
  where $\mathbf{\hat{z}}^{\mathcal T_k}\in \mathbb{R}^{N}$. $N=\sum_{k=1}^{T}N^{\mathcal T_k}$. $\mathbf{W}_{L}$ is a learnable parameter and it is shared by all tasks.  $\mathbf{z}^{\mathcal T_k} \in \mathbb{R}^N$ is the extension of $\mathbf{y}^{\mathcal T_k}$. Different task's label is corresponding with the different part in the vector and the part does not  belong to the task is padded with 0.

\begin{figure}[htbp] 
    \centering 
    \includegraphics[width = 0.4\columnwidth]{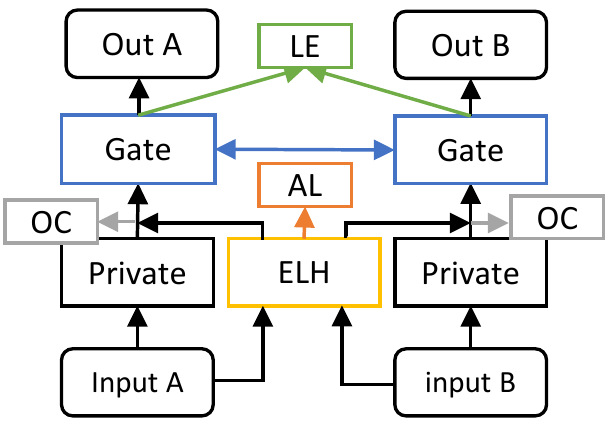} 
    \caption{The ensemble model that integrates all the five MTL algorithms into the basic MLT framework.} 
    \label{fig_Ensemble}
  \end{figure}

To investigate the effectiveness of each MLT method, we integrate all the five methods (i.e., ELH, OC, AL, Gate, LE) into the basic MTL framework illustrated in \ref{fig_base}. The ensemble method is shown in Figure \ref{fig_Ensemble}. We conduct ablation test by removing one of the five methods at each time. Next, we will introduce the experimental setup and analyze the experimental results in detail.

\section{Experimental Setup}
\label{sec:setup}
Since the correlation of data sets has a significant impact on the effect of multi-task learning, we selected a set of high-relevant data sets and a set of low-relevant data sets for multi-task learning experiments.
In order to systematically compare the effect of each MTL method, we conduct extensive experiments on three representative NLP tasks: text classification, text similarity, and textual inference. High-relevant dataset only contains text classification tasks. For low-relevant dataset, Each task contains two benchmark datasets. For fair comparison, in our experiments, we retain the base structure of the five methods and remove all training tricks and extra structures in original works.
 
For the single-task method, we use the architecture introduced in Section 4, in which 1-layer BiLSTM is used for single sentence modeling and 2-layer BiLSTMs with attention are employed for sentence pair modeling. 

It is noteworthy that the share part of diff and adversarial are lighter in the paper comparing linguistic hierarchie, we use a single-layer bi-LSTM in share part when diff and adversarial are used alone or combining with the other component without exploring linguistic hierarchies.
  
\subsection{Tasks and Datasets}
We use the following datasets for our experiments, which are from high-relevant and low-relevant tasks.  Table \ref{tab_data} shows the statistics of these datasets.  

\begin{table}[]
\centering
\small
\begin{tabular}{l|lcc}
\hline
               & Dataset     & Training                     &Testing                      \\ \hline
               & Books       & 1398                         & 400                         \\
               & Electronics & 1398     & 400     \\
               & DVDs        & 1400     & 400     \\
high-relevant  & Kitchen     & 1400     & 400     \\
               & Apparel     & 1400     & 400     \\
               & Baby        & 1300     & 400     \\ \hline

               & CoLA        & 8550                         & 1042                        \\
               & SST-2       & 67349                        & 872                         \\
               & MRPC        & 3668                         & 408                         \\
low-relevant & STS-B       & 5749                         & 1500                        \\
               & MNLI        & 392702                       & 9815                        \\
               & QNLI        & 108436                       & 5732                        \\ \hline
\end{tabular}
\caption{Statistics of the experimental datasets.}
\label{tab_data}
\end{table}
\paragraph{Text Classification}

We apply Amazon product reviews from 6 domains for high-relevant tasks, which includes books, DVDs, cameras, etc. These corpora are classified according to the sentiment of positiveness or negativeness. 
For low-relevant tasks, We use the Corpus of Linguistic Acceptability (\textbf{CoLA})  \cite{warstadt2018neural} and  Stanford Sentiment Treebank (\textbf{SST-2}) \cite{socher2013recursive} datasets to investigate the effectiveness of the MTL methods for text classification task. CoLA  predicts whether an English sentence is linguistically acceptable or not. It uses Matthews correlation coefficient \cite{matthews1975comparison} as the evaluation metric.
SST-2 predicts whether the sentiment of sentences is positive or negative.

\paragraph{Text Similarity}
We use the  Microsoft Research Paraphrase Corpus (\textbf{MRPC}) \cite{dolan2005automatically} and  Semantic Textual Similarity Benchmark \textbf{STS-B} \cite{cer2017semeval} for text similarity task.
MRPC consists of sentence pairs with human annotations denoting whether a sentence pair is semantically equivalent to the other in the pair. Accuracy is used as evaluation metric.
STS-B is a collection of sentence pairs which are manually annotated with similarity scores from one to five, indicating how similar the two sentences are. The evaluation metric is the Pearson and Spearman correlation coefficients.
  
\paragraph{Textual Inference}
We conduct experiments for textual inference task on Multi-Genre Natural Language Inference (\textbf{MNLI}) \cite{williams2017broad} and Question-answering NLI (\textbf{QNLI}) \cite{wang2018glue} datasets.
MNLI contains large-scale, crowd-sourced premise-hypothesis pairs for textual entailment (TE). Each premise-hypothesis pair is labeled with a  relation (i.e., entailment, contradiction, or neutral). 
The evaluation metric is accuracy.
QNLI derived from the Stanford Question Answering Dataset \cite{rajpurkar2016squad}, which has been converted to a binary classification task in GLUE \cite{wang2018glue}. 
QNLI consists of query-candidate-answer tuples, each of which is labeled as positive or negative to indicate whether the given tuple contains the correct answer to the query. The evaluation metric is accuracy.

\subsection{Basic Tasks and Datasets}
For the method of exploring linguistic hierarchies, the shared 3-layer BiLSTM also learn three different basic (auxiliary) tasks, including POS tagging, chunking, and dependency parsing.
\paragraph{POS \& Chunking.} 
We use \textbf{CoNLL-2003} \cite{sang2003introduction} dataset (English  part) for the POS and chunking tasks. 
In CoNLL-2003, each word is labeled with three kinds of tags: part-of-speech tag, chunk tag and named entity tag. The evaluation metric for POS and chunking is word-level accuracy.

\paragraph{Dependency Parsing.}
We use the \textit{Wall Street Journal} (\textbf{WSJ}) part of the Penn Treebank \cite{marcus1993building} for dependency parsing. following the standard data split \cite{collins2002discriminative}. Specifically, we use Sections 2-21 for training, Section 22 for testing.
The evaluation metrics are the Unlabeled Attachment Score (UAS).

\subsection{Hyperparameters Setting}
We use 300-dimensional pre-trained GloVe vectors\footnote{\url{https://nlp.stanford.edu/projects/glove/}} to initialize the word embedding.
The out-of-vocabulary words are randomly initialized with a normal distribution with zero mean and one variance. 
For all high-relevant datasets, we use a hidden state 100 dimensions because of smaller data set and simpler tasks.
For all low-relevant datasets, we set the hidden state of all LSTMs to 300 dimensions. 
All global dropout rate is set to 0.5. 
The other hyper-parameters (e.g., learning rate, maximal sentence, mini-batch size) are tuned on the validation data. 

For dependency parsing task when exploring linguistic hierarchies, we set the dimensions of the two MLP layers as 50 and 100 respectively, similar to \cite{kitaev2018constituency}. We set $\lambda_{adv}= 0.05$ and $\lambda_{OC}= 0.01$.

\section{Experimental Results and Analysis}
\label{sec:result}
\subsection{Single-task vs. Multi-task Analysis}
We analyze the effectiveness of single-task and multi-task methods in detail and conduct an in-depth analysis of the experimental results.  For the multi-task methods, we report the results of the individual MTL tasks on the five benchmarks. In addition, we also report the results of combining different MTL tasks. Due to the large number of combinatorial experiments, we only report the representative results and the complete experimental results are shown in the supplementary file.
For convenience and simplicity, we use letters A, B, C, D, E to represent linguistic hierarchies, gate mechanism, label embedding, orthogonality constrains and adversarial learning, respectively. For example, ABCDE stands for the ensemble model (ENS) of combining the five MTL methods. We repeat each experiment three times and report the average results.  
The results are summarized in Table \ref{tab_result_1} and Table \ref{tab_result_2}.

From Table \ref{tab_result_1} and Table \ref{tab_result_2}, we have the following findings: 
(1) All five MTL methods contribute great improvements to the overall performance, and the linguistic hierarchies method has the best comprehensive performance over all datasets.
(2) The effect of the MTL approach may not be affected by the correlation of the data sets.
(3) The improvement of different MTL methods cannot be superimposed. Next, we analyze the experimental results of the five MTL methods in detail. 

\begin{table*}[h]
\small
\centering
\begin{tabular}{c|c|c|c|c|c|c|c|c}
\hline
Method & \textbf{App.} & \textbf{Kitchen} & \textbf{Books} & \textbf{Elec.} & \textbf{DVDs} & \textbf{Baby} & \textbf{Ave.} & \textbf{Improvement} \\ \hline
Single & 87.50 & 84.75 & 81.75 & 82.50 & 83.00 & 84.00 & 83.92 & -- \\ \hline
A & 89.75 & 87.25 & 86.83 & 88.25 & 82.75 & 88.83 & 87.28 & 3.36 \\ 
B & 88.17 & 85.42 & 83.58 & 83.75 & 83.92 & 86.67 & 85.25 & 1.33 \\ 
C & 88.67 & 85.08 & 83.42 & 83.25 & 83.00 & 86.50 & 84.99 & 1.07 \\ 
D & 89.67 & 87.33 & 86.17 & 86.25 & 82.58 & 88.42 & 86.74 & 2.82 \\ 
E & 89.92 & 87.00 & 85.58 & 87.50 & 83.50 & 88.83 & 87.06 & 3.14 \\ \hline
AB & 89.67 & \textbf{87.92} & 86.58 & 87.83 & 82.83 & 88.83 & 87.28 & 3.36 \\ 
AC & 90.58 & 86.92 & 87.33 & 87.58 & 83.00 & 88.83 & 87.37 & 3.46 \\ 
AD & 89.67 & 87.00 & 87.50 & 87.33 & 82.75 & 88.42 & 87.11 & 3.19 \\ 
AE & 90.33 & 87.67 & 86.50 & \textbf{88.50} & 82.75 & 88.58 & 87.39 & 3.47 \\ 
BC & 88.92 & 85.42 & 83.58 & 84.33 & 82.42 & 85.75 & 85.07 & 1.15 \\ 
BD & 90.00 & 87.17 & 85.92 & 87.42 & 83.67 & 88.75 & 87.16 & 3.24 \\ 
BE & 89.17 & 87.08 & 85.67 & 87.25 & \textbf{84.00} & 88.58 & 86.96 & 3.04 \\ 
CD & 89.00 & 86.50 & 85.17 & 86.75 & 83.83 & 89.33 & 86.76 & 2.85 \\ 
CE & 89.50 & 87.00 & 85.17 & 88.00 & 83.50 & \textbf{88.92} & 87.02 & 3.1 \\ 
DE & 89.92 & 87.50 & 85.83 & 86.92 & 83.50 & 88.25 & 86.99 & 3.07 \\ \hline
ABC & \textbf{91.00} & 87.38 & 87.38 & 87.75 & 82.75 & 88.62 & \textbf{87.48} & \textbf{3.56} \\ \hline
ABCD & 89.83 & 87.25 & \textbf{87.58} & 88.08 & 83.08 & 89.00 & 87.47 & 3.55 \\ \hline
ABCDE & 90.75 & 87.08 & 86.75 & 87.42 & 83.17 & 89.00 & 87.36 & 3.44 \\ \hline

\end{tabular}
\caption{Experiment result on high-relevant datasets. Here, A, B, C, D, E represent linguistic hierarchies, gate mechanism, label embedding, orthogonality constrains and adversarial learning, respectively.}
\label{tab_result_1}
\end{table*}

\begin{table*}[h]
\small
\centering
\begin{tabular}{c|c|c|c|c|c|c|c|c}
\hline
Method& \textbf{SST} & \textbf{CoLA} & \textbf{MNLI} & \textbf{MRPC} & \textbf{QNLI} & \textbf{STS-B} & \textbf{Ave.} & \textbf{Improvement} \\ \hline
Single & 86.35 & 69.19 & 69.13 & 72.30 & 70.83 & 67.47 & 72.55 & -- \\ \hline
A & 89.03 & \textbf{70.5} & 72.93 & \textbf{80.88} & 73.08 & 71.74 & 76.36 & 3.81 \\ 
B & 87.19 & 69.58 & 70.89 & 73.53 & 71.38 & 69.20 & 73.63 & 1.08 \\ 
C & 87.88 & 69.51 & 70.34 & 75.41 & 70.32 & 70.19 & 73.94 & 1.39 \\ 
D & 88.34 & 69.39 & 71.80 & 73.45 & 72.03 & 66.03 & 73.51 & 0.96 \\ 
E & 87.84 & 69.29 & 72.26 & 76.72 & 70.89 & 71.89 & 74.81 & 2.26 \\ \hline
AB & 88.95 & 69.93 & 73.01 & 80.39 & 73.67 & 72.12 & 76.35 & 3.80 \\ 
AC & 89.33 & 70.76 & 72.95 & 80.15 & 73.24 & 72.05 & 76.41 & 3.86 \\ 
AD & 89.03 & 69.86 & 72.75 & 80.07 & 73.71 & 68.25 & 75.61 & 3.06 \\ 
AE & 89.28 & 70.40 & 72.83 & 80.64 & 73.25 & 72.05 & 76.41 & 3.86 \\ 
BC & 87.54 & 69.55 & 70.83 & 73.12 & 71.16 & 70.81 & 73.83 & 1.28 \\ 
BD & 88.26 & 69.32 & 71.31 & 73.77 & 71.88 & 63.66 & 73.03 & 0.48 \\ 
BE & 87.77 & 69.67 & 72.25 & 75.57 & 71.80 & 71.38 & 74.74 & 2.19 \\ 
CD & 87.92 & 69.26 & 70.64 & 73.45 & 71.81 & 64.38 & 72.91 & 0.36 \\ 
CE & 88.07 & 69.58 & 72.30 & 76.59 & 71.51 & 71.96 & 75.00 & 2.45 \\ 
DE & 88.42 & 69.35 & 71.43 & 72.95 & 71.72 & 65.26 & 73.19 & 0.64 \\ \hline
ABC & 89.34 & 70.40 & 73.47 & 79.62 & \textbf{74.66} & \textbf{73.38} & \textbf{76.90} & \textbf{4.35}\\ \hline
ABCD & \textbf{89.45} & 69.93 & 73.45 & 78.76 & 74.39 & 67.96 & 75.59 & 3.04 \\ \hline
ABCDE & 89.14 & 69.96 & \textbf{73.55} & 79.05 & 74.48 & 67.82 & 75.86 & 3.31 \\ \hline

\end{tabular}
\caption{Experiment result on low-relevant datasets.}
\label{tab_result_2}
\end{table*}

\begin{table*}
\small
\centering
\begin{tabular}{l|l|l|l|l|l|l|l}
\hline
& \textbf{App.} & \textbf{Kitchen} & \textbf{Books} & \textbf{Elec.} & \textbf{DVDs} & \textbf{Baby} & \textbf{Ave.}\\ \hline
Single & 87.5 & 84 & 81.75 & 82.5 & 83 & 84.75 & 83.92 \\ \hline
SLH & 87.75 & 82.25 & 84.25 & 84.00 & 82.75 & 85.25 & 84.37 \\ \hline
No Auxiliary Tasks & 89.83 & 87.58 & 85.33 & 87.33 & 83.41 & 87.91 & 86.9 \\ \hline
Pos & 90.50  & 87.25  & 86.88  & 87.13  & 82.75  & 89.00  & 87.25 \\ 
Chunking & 90.13  & 87.13  & 85.63  & 86.88  & 82.88  & 88.88  & 86.92 \\ 
Parsing & 90.63  & 87.25  & \textbf{87.75}  & 87.50  & 85.25  & 88.75  & \textbf{87.85} \\ \hline
Pos+Chunk & \textbf{90.75}  & 87.75  & 85.75  & \textbf{87.75}  & 83.00  & \textbf{90.75}  & 87.63 \\ 
Pos+Parsing  & \textbf{90.75}  & \textbf{88.38}  & 86.50  & 87.75  & \textbf{83.63}  & 88.88  & 87.65 \\ 
Chunk+Parsing  & 90.38  & 87.88  & 86.00  & 87.00  & 82.75  & 89.13  & 87.19 \\ \hline
ELH  & 89.75 & 87.25 & 86.83 & 88.25 & 82.75 & 88.83 & 87.28 \\ \hline

\hline
\end{tabular}
\caption{Exploring linguistic hierarchies method with Partial auxiliary tasks result on high-relevant datasets. SLH stands for Single task with Linguistic Hierarchies}
\label{partial_auxiliary_tasks_1}
\end{table*}

\begin{table*}
\small
\centering
\begin{tabular}{l|l|l|l|l|l|l|l}
\hline
 & \textbf{SST} & \textbf{CoLA} & \textbf{MNLI} & \textbf{MRPC}  & \textbf{QNLI} & \textbf{STS-B} & \textbf{Ave.} \\ \hline
Single & 86.35 & 69.19 & 69.13 & 72.3 & 70.83 & 67.47 & 72.58 \\   \hline
SLH & 88.45 & 69.87 & 72.69 & 77.53 & 72.76 & 66.46 & 74.63 \\  \hline
No Auxiliary Tasks & 88.15 & 69.61 & 70.78 & 74.02 & 70.88 & 73.88 & 74.55 \\  \hline
Pos & 88.45 & 69.74 & 72.75 & 78.43 & 71.68 & 74.11 & 75.86 \\
Chunking  & 88.65 & 69.45 & 73.19 & 78.02 & 72.47 & \textbf{74.17} & 75.99 \\
Parsing & 88.65 & 69.99 & \textbf{73.60} & 79.49 & 73.04 & 72.6 & 76.23 \\ \hline
Pos+Chunk & 88.57 & 70.19 & 72.68 & 79.98 & 71.98 & 73.15 & 76.09 \\
Pos+Parsing & 88.95 & 69.93 & 73.1 & 81.05 & 73.3 & 72.08 & 76.4 \\
Chunk+Parsing & 88.95 & 70.22 & 73.22 & \textbf{81.86} & \textbf{73.75} & 72.78 & \textbf{76.80} \\ \hline
ELH & \textbf{89.03} & \textbf{70.50} & 72.93 & 80.88 & 73.08 & 71.74 & 76.36 \\ \hline\end{tabular}
\caption{Exploring linguistic hierarchies method with Partial auxiliary tasks result on low-relevant datasets.}
\label{partial_auxiliary_tasks_2}
\end{table*}

\paragraph{Linguistic hierarchies}
From the results, we can observe that exploring linguistic hierarchies has the best overall performance among the compared individual MTL techniques on both two types of datasets.
This verifies that the auxiliary basic tasks (i.e., POS tagging, chunking, and dependency parsing) with shared parameters can capture more comprehensive word-level and sentence-level semantic features, thereby yielding the strong performance of three representative NLP tasks. In addition, adding auxiliary basic tasks increases the amount of training data and therefore increases the generalization of the model. 

\paragraph{Gate mechanism}
Gate mechanism achieves better results than the single-task model on two data sets.
Gate mechanism has an ability to select potentially useful features, which can reduce the interference among tasks \cite{hochreiter1997long}.
\paragraph{Label embedding}
Surprisingly, label embedding outperforms the single-task model significantly with its lightweight structure, only with this component the average performance has improved 1.07 and 1.36 on two datasets. This may be because that label embedding method can leverage label information of each task by mapping labels into dense, low-dimension and real-value vectors with semantic implications, which captures the semantic correlations among tasks.
\paragraph{Orthogonality constraint}
Comparing with the result of single-task, the average performance of orthogonality constraint increases 2.82 and 0.93 on average respectively. We can see that the structure has significantly higher gains for high-relevance tasks than for low-relevance tasks. 

We believe that although orthogonality constraint can make a difference between the shared and private feature learned by shared part and private part respectively, bu In the case of low-relevance, the orthogonality constraint in share layer may not be able to obtain enough shared information, resulting in less task promotion.
\paragraph{Adversarial learning}
The adversarial learning framework has the second largest increase in the single method. 
The simple adversarial learning method can lead to a good shared feature space contain more common information and no task-specific information, and it works well on different data sets.

\paragraph{Summary}

After analyzing the performance of the five MTL methods over the single-task model, we find that linguistic hierarchical information performs better than other individual MTL methods.
The gate mechanism introduces a large number of parameters, but the effect improvement is not significantly better others.
Label embedding has a simple structure and performs better than some more complex MTL methods.
Orthogonality constraint are highly correlated with specific datasets and have different influence on results.
Adversarial learning can improve the performance of overall datasets without introducing extraneous data and adding a large number of parameters.

\begin{table*}[h]
\small
\centering
\begin{tabular}{l|l|l|l|l|l|l|l|l|l|l}
\hline
&\textbf{POS} & \textbf{CHUNK} & \textbf{DP} & \textbf{App.} & \textbf{Kitchen} & \textbf{Books} & \textbf{Elec.} & \textbf{DVDs} & \textbf{Baby} & \textbf{Ave.}  \\ \hline
ENS  & 93.70  & 95.33  & 92.02  & 90.75  & 87.08  & 86.75  & 87.42  & 83.17  & 89.00  & 87.36 \\
ENS-JMT  & 93.34 & 95.07 & 91.61 & 91.31 & 89.31 & 86.33 & 89.03 & 84.11 & 90.19 & 88.38 \\ \hline
\end{tabular}
\caption{Experiment result on ensemble model  with joint many-task model (ENS-JMT) and our ensemble model (ENS) on high-relevant datasets.}
\label{tab_result_JMT_1}
\end{table*}

\begin{table*}[h]
\small
\centering
\begin{tabular}{l|l|l|l|l|l|l|l|l|l|l}
\hline
& \textbf{POS} & \textbf{CHUNK} & \textbf{DP} & \textbf{SST} & \textbf{CoLA} & \textbf{MNLI} & \textbf{MRPC} & \textbf{QNLI} & \textbf{STS-B} & \textbf{Ave.}  \\ \hline
ENS  & 94.11 & 95.65 & 92.48 & 89.14 & 69.96 & 73.55 & 79.05 & 74.48 & 67.82 & 75.86\\
ENS-JMT    & 93.83  & 95.11  & 92.1 & 88.96 & 70.06 & 73.24 & 78.92 & 74.05  & 69.86  & 75.84  \\ \hline
\end{tabular}
\caption{Experiment result on ensemble model (ENS)  with joint many-task Model (ENS-JMT) and our ensemble model (ENS) on low-relevant datasets.}
\label{tab_result_JMT_2}
\end{table*}




\subsection{In-depth Analysis of Exploring Linguistic Hierarchies}
As shown in Table \ref{tab_result_1} and Table \ref{tab_result_2}, we can observe that exploring linguistic hierarchies achieves the best performance among the individual MTL methods on most datasets. To investigate this unexpected observation, we conduct ablation test to analyze the efficacy of each basic task (i.e., POS tagging, Chunking, dependency parsing) to the performance of ELH method by discarding one or two auxiliary basic tasks each time. In addition, we also compared single task trained with hierarchical language structures (SLH) to verify that the increment of this structure comes from auxiliary tasks or multitasking. Result are shown in Table \ref{partial_auxiliary_tasks_1} and Table \ref{partial_auxiliary_tasks_2}.

Combining all the auxiliary tasks does not achieve the best results on average score, and high-level auxiliary tasks can bring better results. By observing these tables, best performance for all missions only contains dependency parsing on high-relevant datasets, and on low-relevant datasets the model with the best performance only include chunking and dependency parsing.

We also found that the key to the performance improvement of exploring linguistic hierarchies is the auxiliary task in the low-relevance data sets, while in the high-relevant data sets, the hard parameter sharing by multi-task learning has a more obvious effect. In other words, In the case of simple hard sharing with out auxiliary tasks, multi-task learning with high data set correlation can bring more improvements than low-relevance datasets.
Comparing the performance on exploring linguistic hierarchies with no auxiliary tasks and the SLH model, we can see that in the high-relevant data set, exploring linguistic hierarchies with no auxiliary tasks outperform SLH by 2.53 on average. But on the low-relevance tasks, the SLH method performs better than exploring linguistic hierarchies with no auxiliary tasks on all the dataset except STSB, in other words, the introduction of language hierarchical method is significantly better than multi-task learning in these datasets.
We believe that the structure can capture the benefits of both hierarchical language structure and multi-task learning at the same time, which is why the structure can stand out in many multi-task learning methods.

It is noteworthy that the implementation of exploring linguistic hierarchies, denoted as joint multi-task  model (JMT), in the original paper \cite{sogaard2016deep} is slightly different from the implementation in this paper. In \cite{sogaard2016deep}, the loss functions to different layers are injected from bottom to top layers, and the amount of upgrade on the lower layer is limited. However, in this paper, we add the losses of all layers and optimizes them at the same time.  
Since there is no open source code, we also recurring the paper's method and follow the hyper-parameters in the paper. We compared the performance of the ensemble model and the ensemble model replacing JMT training method on linguistic hierarchies. The results are reported in Table \ref{tab_result_JMT_1} and Table \ref{tab_result_JMT_2}.
From the results, we can observe that our training strategy performs slightly better than \cite{sogaard2016deep} on the auxiliary tasks. 

\subsection{Overfitting Analysis}
\begin{figure}[htbp!]  
\centering 
\includegraphics[width = 0.9\columnwidth]{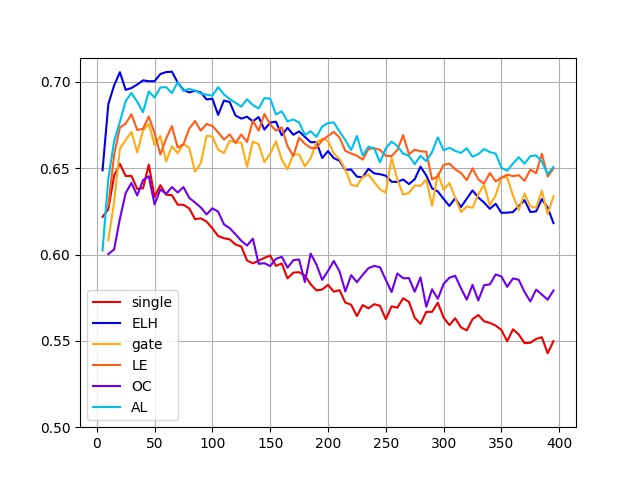} 
\caption{Overfitting on STS-B(Taking an average of every five values)} 
\label{learning-curve}
\vspace{-0.3cm}
\end{figure}

Multi-task learning is supposed to alleviate the overfitting problem by leveraging useful information contained in multiple related tasks to improve the generalization performance of all the tasks. Based on our empirical observation, there is high-risk of overfitting on STS-B dataset.  
To investigate the performance of the five MTL methods on alleviating overfitting problem, we illustrate the learning curves of these models in Figure \ref{learning-curve}, where each method iterates over more than 400 epochs. 
Compared to the highest value, single-task model decreased by 18.8\%, while ELH, Gate, Label Embedding, Orthogonality Constraints and Adversarial Learning decreased by 14.21\%, 10.22\%, 9.35\%, 13.17\% and 9.18\% respectively.

From Figure \ref{learning-curve}, we can observe that all MTL methods can alleviate the overfitting problem. In general, the more parameters of MTL model, the worse its effect on overfitting. In particular, the adversarial learning and label embedding methods greatly reduce the degree of overfitting. We think this is because the method uses the least amount of parameters to achieve the interaction of multiple tasks. The OC method with small amount of parameters doesn't behave well on overfitting problem, we think because of the addition of a LSTM layer.

\section{Conclusion}
\label{sec:conclusion}
There is no existing systematic exploration and comparison of different MTL architectures and learning mechanisms for their strong performance in-depth.
To the best our knowledge, this work is the first to investigate the effects of five widely used MTL methods with deep neural networks for a broad range of representative NLP tasks including text classification, semantic textual similarity, natural language inference, question answering, part-of-speech tagging, and chunking. 
By identifying the essential MTL architecture designs that contribute significantly to the success of MTL, we propose a hybrid architecture built with different components borrowed from previous MTL methods. 
Finally, we present a series of findings that (i) all six individual MTL methods contribute great improvements to the overall performance on the experimental datasets. In addition, the improvements of the MTL methods cannot be superimposed; (ii) using linguistic hierarchical information performs better than other individual MTL methods. However, using all the three auxiliary basic tasks does not achieve the best results. The upper-level auxiliary task brings more improvement than the lower-level auxiliary tasks;
(iii) combining linguistic hierarchies, gate mechanism, and label embedding methods  can achieve best results on most of the datasets. 

In the future, we would like to investigate the effectiveness of these MTL methods in alleviating the negative knowledge transfer problems among different tasks, especially when the optimal solutions of the tasks are quite different.  In addition, we also plan to devise a more effective MTL framework when considering the strengths and weakness of existing MTL methods for alleviating the negative knowledge transfer problems.

\section*{Conflict of Interest Statement}
The authors declare that they have no known competing financial interests or personal relationships that could have appeared to influence the work reported in this paper.




\end{document}